\documentclass{article}

\usepackage{authblk}
\usepackage{graphicx} % Required for inserting images
\usepackage{algorithm}
\usepackage{algpseudocode}
\usepackage{stmaryrd}
\usepackage{float} % here for H placement parameter\
\usepackage[french,english]{babel}
\usepackage{hyperref}
\usepackage{multirow}%
\usepackage{amsmath,amssymb,amsfonts}%
\usepackage{amsthm}%
\usepackage{mathrsfs}%
\usepackage[title]{appendix}%
\usepackage{xcolor}%
\usepackage{textcomp}%
\usepackage{manyfoot}%
\usepackage{booktabs}%
\usepackage{algorithm}%
\usepackage{listings}%
\usepackage[T1]{fontenc} 

\begin{document}

%\title[Article Title]{A Survey on Recent Advances in Self-Organizing Maps}
\title{A Survey on Recent Advances in Self-Organizing Maps}


\author[1,2]{Axel Guérin}
\author[1]{Pierre Chauvet}
\author[2]{Frédéric Saubion\thanks{corresponding author frederic.saubion@univ-angers.fr}}

\affil[1]{Univ Angers,, LARIS, SFR MATHSTIC, F-49000 Angers, France and  Université Catholique de l'Ouest (France)}
\affil[2]{Univ Angers, LERIA, SFR MATHSTIC, F-49000 Angers, France }

\date{} % Pas de date si vous ne souhaitez pas en afficher

\maketitle

\begin{abstract}
Self-organising maps are a powerful tool for cluster analysis in a wide range of data contexts. From the pioneer work of Kohonen, many variants and improvements have been proposed. This review focuses on the last decade, in order to provide an overview of the main evolution of the seminal SOM algorithm as well as of the methodological developments that have been achieved in order to better fit to various application contexts and users' requirements. We also highlight a specific and important application field that is related to commercial use of SOM, which involves specific data management.
\end{abstract}

\section{Introduction}

The Self-Organising Map algorithm is a well-known approach for unsupervised learning, designed to distill a high-dimensional dataset into a more manageable, typically two-dimensional, representation. Imagine a dataset full of $p$ measured variables across $ n $ observations. A Self-Organising Map elegantly organises similar observations into groups and visually displays them on a map.

This model, also known as Kohonen maps or Kohonen networks, has been introduced by Teuvo Kohonen \cite{kohonen82,Kohonen97}. Unlike conventional neural networks, which rely on error correction, SOM training relies on competitive principles. Kohonen drew inspiration from biological paradigms, in particular the neural models \cite{minsky1969} and Alan Turing's pioneering theories of morphogenesis \cite{turing52}. Basically, self-organising maps serve as powerful tools for dissecting and visualising complex data landscapes, facilitating a deeper understanding of the intricate structures and relationships that permeate multidimensional datasets.

Self-organising maps, like most artificial neural network architectures, operate in two distinct modes: training and mapping. In the training phase, a set of input data (representing the 'input space') is used to generate a reduced dimensional representation of the original data (defining the 'map space'). Subsequently, in the mapping phase, additional input data is classified using the generated map.

In most scenarios, the training objective is to project a $p$-dimensional input space onto a two-dimensional map space. The map space is structured into units called {\rm nodes} or {\em neurons}, which are arranged in a two-dimensional hexagonal or rectangular grid. The number and arrangement of nodes is determined in advance, depending on the overall goals of data analysis and exploration.

Each node within the map space is associated with a weight vector, which reflects the node's position in the input space. Although the nodes within the map space remain fixed, the training phase involves moving the weight vectors towards the input data (minimizing a distance measure such as Euclidean distance), while preserving the topology that derives from the structure of the map space. Once training is complete, the map can classify additional observations in the input space by identifying the node whose weight vector is closest (by the smallest distance measure) to the vector associated with the observation in question.

While being rather simple, SOMs are used in a wide variety of domains. This broad scope includes visualisation, feature map generation, pattern recognition and classification. In his 1997's book\cite{Kohonen97}, Kohonen identified different applications, including machine vision, image analysis, optical character recognition, script reading, speech analysis, acoustic and musical studies, signal processing, radar measurements, telecommunications, industrial measurements, process control, robotics, chemistry, physics, electronic circuit design, non-imaging medical applications, linguistic and AI problems, mathematical puzzles, and neurophysiological research. 

Let us mention recent applications in economy to identify the determinants of the debt crisis \cite{AllegretC23}, in software engineering for classifying Model-View-Controllers \cite{GuamanDP21}, in linguistic for sentiment classification \cite{TacheGI21}, in combinatorial optimization for planning patrols of indoor unmanned aerial vehicule by means of TSP solving \cite{Frezza-Buet20}, in medicine for epidemiology study \cite{SilvaSCC18}, in image processing including efficient dedicated implementation \cite{SousaPD20}. This list highlights the current relevance of SOM methodology across an array of diverse fields of application.

The purpose of this survey is to provide an overview of the recent advancements in SOM-based algorithms over the past decade. As previously mentioned, SOMs remain widely used for data analysis across various domains. However, like many traditional machine learning techniques, the SOM algorithm is computationally intensive and needs adaptation for different use cases. Recent research has focused on enhancing the efficiency and applicability of SOM algorithms. These improvements span various stages of the general method, from preprocessing input datasets to fine-tuning the algorithm's components and parameters, and ultimately providing optimal data visualizations for users.

Our survey aims to highlight representative works that showcase the innovations in this field. We follow a methodological guideline to organize these works along the entire map generation process. Given the broad range of application fields, we have chosen to focus on the use of SOMs in a commercial context, based on our own experience in this domain.

\section{The SOM Algorithm in Brief}

Let us first recall the basic principle of the SOM algorithm  that involves the following key steps:

\begin{enumerate}
    \item \textbf{Initialization}: Initialize the weight vectors of the neurons, typically with small random values or by sampling from the input data distribution.
    \item \textbf{Training}: For each input vector, repeat the following steps for a fixed number of iterations or until convergence:
    \begin{enumerate}
        \item \textbf{Best Matching Unit (BMU) Search}: Find the neuron with the weight vector that is closest to the input vector (using a distance measure such as Euclidean distance).
        \item \textbf{Weight Update}: Adjust the weights of the BMU and its neighboring neurons to move closer to the input vector. The degree of adjustment decreases with time.
    \end{enumerate}
    \item \textbf{Convergence}: After sufficient iterations, the weight vectors of the neurons stabilize, forming a topologically ordered map that reflects the input data structure.
\end{enumerate}

\begin{algorithm}[htbt]
\begin{algorithmic}
    \State Initialisation
    \State $ \epsilon_0 \gets $ initial learning rate
    \State $ \sigma_0 \gets $ initial radius
    \State $ T_{max} \gets $ maximum number of iterations
    \State $ X \gets $ input data matrix ($n$ observations using $p$ variables/characteristics)
    \State $ d \gets $ distance between two vectors
    \State $ d^* \gets $ distance between two polygons on the map (depends on topology)
    \State $ E \gets $ weight map, $ E = \{ W_i, i \in \llbracket 1, 4 \sigma^{2}_{0} \rrbracket \}$, initialized randomly
    \For{$ t \gets 1 $ \textbf{to} $ T_{max} $}
        \State Select $ X^* $ randomly from $ X $
        \State Find $ W^* $, the closest node on the map to $ X^* $
        \State $ W^* = \arg\min(d(W, X^*), W \in E) $
        \State Update the weights of the neurons neighboring $ W^* $
        \For{$ i \gets 1 $ \textbf{to} $ 4 \sigma^{2}_{0} $}
            \State $ W_i(t + 1) \gets W_i(t) + \Theta(t) \epsilon(t)(X^* - W_i(t)) $, with:
            \State $ \Theta(t) = e^{-d^*(W^*, W_i) / 2 \sigma^2(t)} $
            \State $ \sigma(t) = \sigma_0 e^{-t / T_{max}} $
            \State $ \epsilon(t) = \epsilon_0 e^{-t / T_{max}} $
        \EndFor
    \EndFor
\end{algorithmic}
\caption{Simple SOM Algorithm}
\label{algo:som}
\end{algorithm}

The result of this algorithm is the weight map $E$ containing weights of dimensions $p$ corresponding to the $p$ input variables. Each input element can be associated with the nearest weight. These $p$ variables can be represented on a grid of squares, hexagons or triangles (depending on the topology defined). Each element of the grid is assigned a colour according to its value.

The map of weights constitutes a topological space (generally a discrete topology, i.e. any subset of the map E is an open). Different geometric shapes can be used to represent the neural network. The 3 most common shapes produce a pavement with regular shapes, i.e. triangles, squares or hexagons. This topology space has a distance corresponding to the minimum number of polygons to be covered between two points. This distance is denoted $ d* $ in Algorithm \ref{algo:som}.

\begin{figure}[H]
    \centering
    \includegraphics[width=0.5\textwidth]{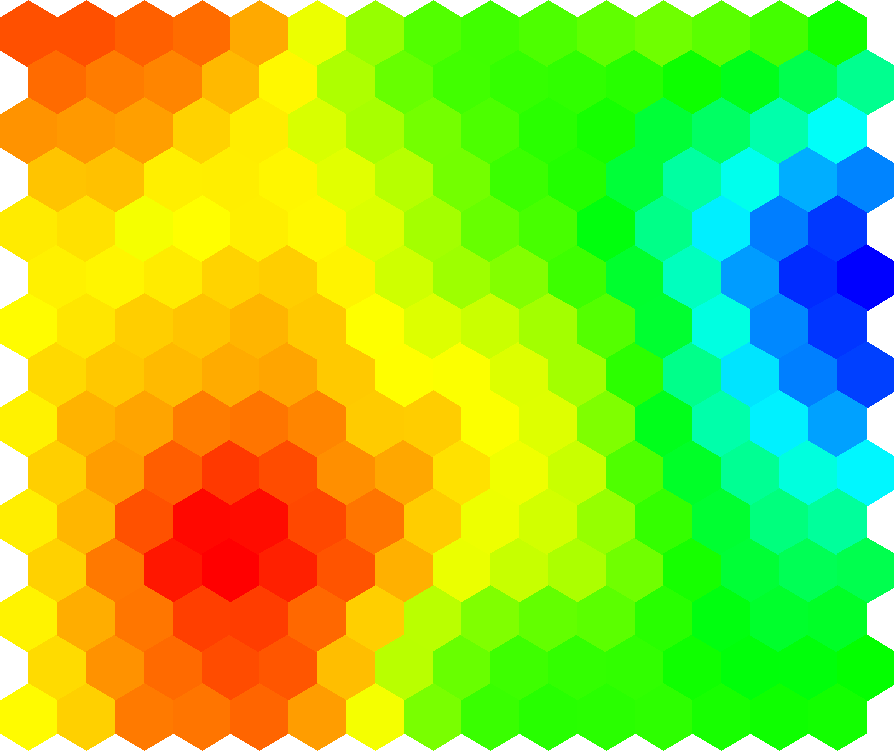}
    \caption{Map of hexagons representing 1 variable from a dataset of 417 entries}
    \label{fig:map-sample_hexagone_15x15}
\end{figure}

Figure \ref{fig:map-sample_hexagone_15x15} is a sample of the result obtained by the algorithm. 

\medskip
\noindent
SOM are traditionally designed as two-dimensional maps, providing a visually intuitive way of representing complex data. However, SOMs can be extended to higher dimensions, such as three dimensions. This extension increases the map's ability to model more complex relationships between data, but it also presents significant challenges. When the number of dimensions of a SOM map is increased to three, visual appearance and interpretability can become more complicated \cite{Zin2014,Gorricha2011}. Although three-dimensional maps provide a richer representation of data, they lose visual clarity, making analysis and interpretation less straightforward. Moreover, visualising such maps often requires specific tools and techniques to explore three-dimensional space effectively.

Furthermore, the number of neurons in a SOM map increases exponentially with the addition of dimensions. In a three-dimensional map, this phenomenon translates into a rapid increase in the number of neurons required, which can lead to computational challenges and increased complexity in model training and optimisation. When it comes to tiling in a three-dimensional map, the structuring of space becomes more complex. Using cubes as the basic units for tiling is a common approach, as it provides a structured and systematic way of dividing three-dimensional space. However, this method simplifies the representation of the data space and may not fully capture the complexity inherent in some data distributions.

Interactive visualization \cite{Schreck10} remains a challenge and could enhance the analytical experience, offering real-time manipulation, zooming, rotation, and dynamic filtering, enriching data exploration. Yet, developing such tools demands substantial investment in time, resources, and specialized expertise, considering interface design, advanced coding, performance optimization, and accessibility. While 3D SOMs offer undeniable advantages in modeling three-dimensional structures, careful consideration of computation time and visualization challenges is crucial when selecting the SOM configuration for specific data analysis tasks.

\section{Evolution of the SOM Algorithms During the Last Decade}

\begin{figure}[H]
    \centering
    \includegraphics[width = \textwidth]{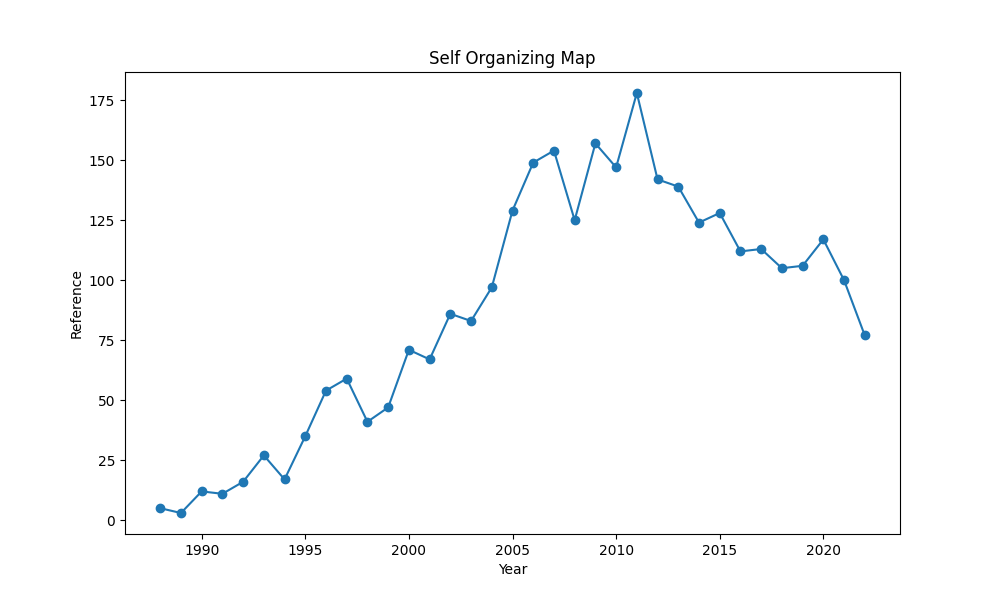}
    \caption{Evolution of the number of references per year for the ‘Self Organizing Map’ search on the DBLP database (\url{https://dblp.org/})}
    \label{fig:review-self-organizing-map}
\end{figure}

In this section dedicated to the later evolution of Self-Organizing Maps, we will limit ourselves to examining the publications and research carried out over the last ten years. This period, from 2014 to now, represents a phase that remains dynamic and innovative in the development of SOMs as highlighted by Figure \ref{fig:review-self-organizing-map}. 

\begin{figure}[H]
    \centering
    \includegraphics[width = \textwidth]{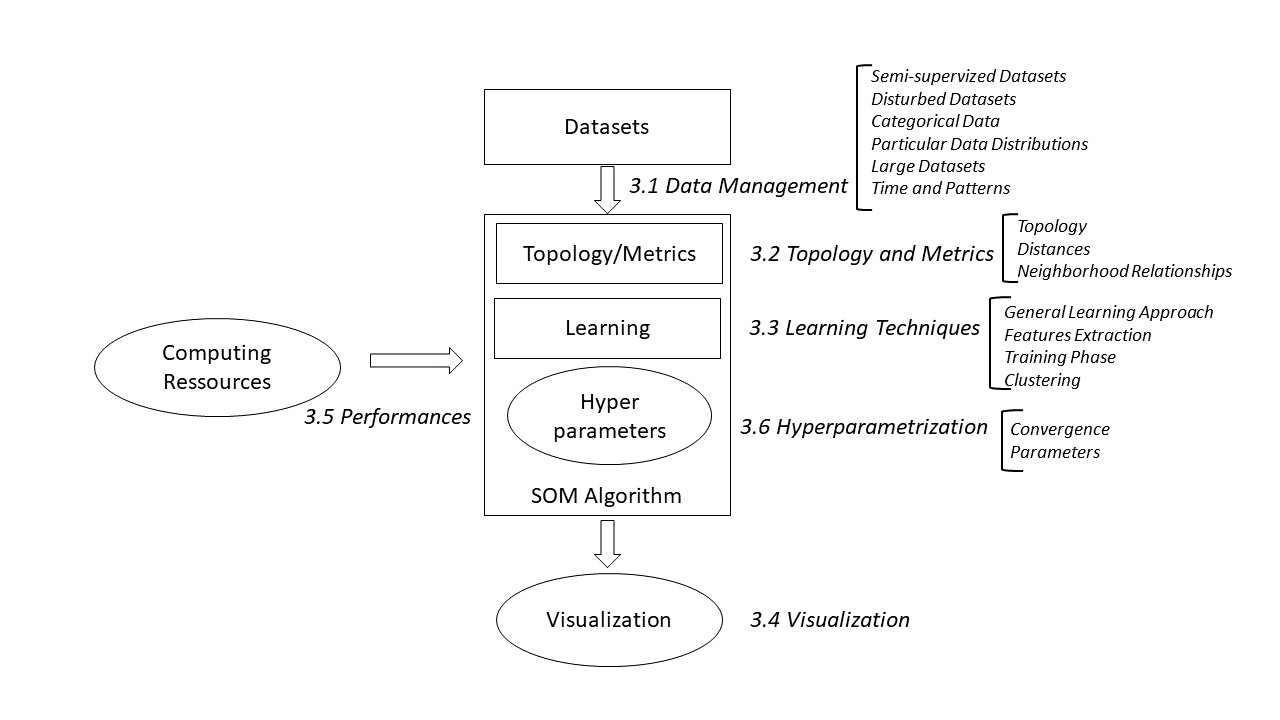}
    \caption{Organization of the survey according to the different components and key features of SOM algorithms}
    \label{fig:orga_general}
\end{figure}

To review some of the most representative works from this period, we have organized our exploration according to four distinct improvement directions for SOM algorithms, as depicted in Figure \ref{fig:orga_general}. This figure details the organization of the next subsections of this review. 

\begin{itemize}
    \item {\bf Data management : } SOM can be used to visualize very different types of data (numerical, categorical...). Moreover, the input data set may present difficulties in various real cases (missing values, heterogeneity..).
    \item {\bf Topology and metrics} SOM algorithms require the selection of a topology to represent the map as well as distances and neighborhood relations for building this map. 
    \item {\bf Learning techniques } : since SOM algorithms belong to machine learning techniques they involve choices concerning the setting of the learning process as well as for the different learning phase, including training and preparation of the data. 
    \item {\bf Visualization : } data visualization is obviously the main goal of the SOM algorithm. Therefore, different techniques and options can be selected to improve the resulting maps
    \item {\bf Performances : } as many machine learning approaches, the SOM algorithm can be very computationally costly. Proposals have been made to enhance the classification process, including dedicated hardware solutions
    \item {\bf Hyperparameters }: Hyperparameterisation refers to the process of selecting and adjusting hyperparameters, which are configuration parameters that influence the behaviour and performance of the SOM  algorithm to enhance its results. 
\end{itemize}

\subsection{Data Management}

The input datasets may present difficulties due to missing or incomplete information (e.g., missing labels). Moreover, specific kinds of data may require dedicated solutions (e.g., data described by statistical distributions). 

\paragraph{Semi-supervized Datasets}

"A Semi-Supervised Self-Organizing Map for Clustering and Classification" \cite{DBLP:conf/ijcnn/BragaB18} presents SS-SOM, an innovative method for managing semi-supervised datasets. Faced with the prevalence of data with few labelled samples, SS-SOM offers a flexible solution, alternating between supervised and unsupervised learning depending on the availability of labels. This approach is distinguished by its ability to excel in conditions where labelled data is limited, while remaining effective for fully labelled sets. This makes SS-SOM particularly useful for clustering and classification scenarios in semi-supervised data contexts.

\medskip
In "Constrained Semi-Supervised Growing Self-Organizing Map" (CS2GS) \cite{DBLP:journals/ijon/AllahyarYH15}, the CS2GS algorithm is introduced as a solution for online semi-supervised clustering. This model focuses on the use of constraints in unlabelled samples, a useful strategy when labelled data is sparse or difficult to obtain. CS2GS modifies the online learning of semi-supervised Self-Organizing Maps and adapts it into a constrained metric learning problem solved by iterative Bregman projections.

This approach addresses the challenges posed by massive or streaming datasets, where offline and batch methods are limited. Tests with synthetic and real-world data, including UCI datasets and a bilingual corpus for machine translation, demonstrate the effectiveness of CS2GS in online semi-supervised clustering. The results highlight the benefit of incorporating unlabelled samples to improve the accuracy of the system in realistic data scenarios.

\medskip
{\bf Remark : }Note that CS2GS and SS-SOM focus on improving semi-supervised clustering, but with distinct approaches. CS2GS targets the optimisation of clustering in a context of data flow and constraints on unlabelled data, while SS-SOM focuses on the flexibility of learning in various label availability scenarios, offering versatility for clustering and classification in semi-supervised data contexts.

\paragraph{Disturbed Datasets}

\medskip
"Semi-automated data classification with feature weighted self organizing map"  (FWSOM) \cite{DBLP:conf/icnc/StarkeyA17} introduces the Feature Weighted Self-Organizing Map (FWSOM), a breakthrough in data classification. FWSOM uses topology information from a standard SOM to automatically guide the selection of important entries during training, improving the classification of data with irrelevant entries. This method has demonstrated improved classification accuracy over the standard SOM and other existing classifiers, on both synthetic and real datasets. A notable strength of FWSOM is its ability to identify relevant features, thereby improving the classification performance of other methods. This development represents an important step towards more accurate and automated data classification, particularly in scenarios where filtering out irrelevant features is crucial.

\medskip
\cite{DBLP:journals/simpa/GouveaGMC23} presents IntraSOM, a new Python library dedicated to the implementation of Self-Organizing Maps. The library is distinguished by its ability to support hexagonal grids and toroidal topology, offering greater flexibility in the modelling and analysis of complex data. A major strength of IntraSOM is its effective handling of missing data during the training process, a frequent problem in real datasets. In addition, IntraSOM offers advanced visualisation tools and efficient clustering algorithms, making it a valuable tool for researchers and practitioners wishing to explore and analyse complex datasets intuitively. The library is designed to be accessible and complete, with an extensible framework that makes it easy to integrate with other Python algorithms and libraries.

\paragraph{Categorical Data}

\medskip
In  \cite{DBLP:journals/ida/HsuKJC19}, the authors address an important issue in the processing of categorical data with Self-Organizing Maps. Initially, SOMs were designed to process numerical data, but this study extends their applicability to categorical data by integrating distance hierarchies that reflect the semantic structure of this data.

The integration of learned distance hierarchies with the extended SOM enables better management and visualisation of mixed datasets, including both numerical and categorical elements. Experiments carried out in the study to verify the feasibility and compare the performance of different unsupervised learning methods demonstrate the effectiveness of the approach.

\medskip
A significant contribution for processing categorical data, a common challenge in pattern recognition and data mining has been achieved in \cite{DBLP:journals/asc/CosoFDNRV15}. Traditionally, SOMs are designed to process numerical data, and integrating categorical data often involves converting it into binary codes. However, this transformation can lead to distortion during network training and subsequent data analysis. To overcome this limitation, this paper proposes an innovative SOM architecture that can process categorical values directly without the need for prior conversion into binary codes.

A key aspect of this new architecture is its ability to effectively mix numeric and categorical data, assigning equal weight to all features. This approach ensures a more faithful and balanced representation of the different types of data within the network. In addition, the proposed architecture is described as scalable, suggesting its ability to handle large and diverse datasets.

\medskip
"Hierarchical SOM (hSOM): Visualizing Self-Organizing Maps for Categorical Data" \cite{DBLP:conf/iv/KilgoreTCN20} tackles the challenge of analyzing and visualizing multidimensional data, particularly when categorical variables are involved. The authors introduce an innovative method, hSOM, which integrates a histogram into the traditional SOM visualization to better represent categorical data. This integration enhances the understanding of how categorical data is distributed within the SOM, offering a more intuitive and informative analysis. The proposed approach has wide-ranging applications, making SOMs more accessible and useful in fields like biomedical research, marketing, and social behavior analysis where categorical data analysis is vital.

\paragraph{Particular Data Distributions}

\medskip
"Batch Self-Organizing Maps for Distributional Data with an Automatic Weighting of Variables and Components" (DBSOM) \cite{DBLP:journals/classification/CarvalhoIVB22} introduces  the DBSOM algorithm adapted to variables described by probability or frequency distributions. DBSOM uses the Wasserstein distance $L2$, commonly applied in distributional data analysis, as the loss function. A key innovation lies in the introduction of automatically learned relevance weights for each variable with a distributional value, as well as for the components of the Wasserstein distance, which is broken down into means and the size/shape of the distributions.

This approach makes it possible to emphasise the importance of the different characteristics of the distributions in the value of the distance. The proposed algorithms have been validated on real datasets with distributional values, demonstrating the effectiveness of the DBSOM method for managing and interpreting complex data in a multi-dimensional context.

\medskip
 Smoothed SOM (S-SOM) \cite{DBLP:journals/isci/DUrsoGM20} is designed to offer increased robustness to the presence of outliers or atypical data. This development aims to overcome a common limitation of traditional SOMs, which can be sensitive to these outliers, affecting the quality of input density mapping, vector quantization and clustering.

The key innovation of the S-SOM lies in the modification of the learning rule. This modification smoothes the representation of outlier input vectors on the map, using a complementary exponential distance between the input vector and its closest codebook. This approach enables the S-SOM to better handle atypical data and produce a map that is more representative of the real structure of the data.

\paragraph{Large Datasets}

\cite{DBLP:journals/ijcia/CordelA18} introduces an innovative method for the rapid visualization of large amounts of data. The Self-Organizing Map is known for its ability to efficiently cluster and visualize data in a lower dimensional space, but its computational complexity of $o(n^2)$ makes it less suitable for large datasets. To overcome this challenge, the authors propose a force-directed visualisation method that mimics the capabilities of SOMs, while significantly reducing complexity to $o(n)$. The approach is based on force-driven fine-tuning of the 2D data representation. To demonstrate the effectiveness and potential of this method as a rapid visualisation tool, it is applied to the 2D projection of the MNIST handwritten digits dataset. This work suggests a promising advance for applications requiring rapid and accurate data synthesis and visualisation, such as trade trends, disaster response and disease outbreaks.

\cite{DBLP:conf/ijcnn/AntoninoA16} addresses a crucial problem in complex data processing: the management of high-dimensional data, often characteristic of various fields such as bioinformatics, medical imaging, marketing research or social network analysis.

The authors highlight the fact that traditional clustering algorithms often see their performance decline as data dimensions increase. To overcome this challenge, they propose a novel multi-view clustering approach based on a self-organising map that is adaptive and scalable over time. This method is distinguished by its ability to efficiently process real data characterised by high sparsity and dimensionality, as well as by diverse representations.
Their innovative solution uses a subspace clustering approach, dynamically adapting to different ‘views’ or perspectives of the data. Experimental results show that their model outperforms other state-of-the-art models specifically designed for multi-view clustering.

\paragraph{Time and Patterns}

Innovative schemes for automated and weighted self-organizing time maps (SOTMs) are proposed in \cite{DBLP:journals/kais/Sarlin15}. These maps provide a visual method for evolutionary clustering, which consists of producing a sequence of clustering solutions over time, a process called visual dynamic clustering.

Automating SOTMs involves data-driven parameterisation, as well as adapting the training to the characteristics of the data at each time step. The objective of weighted SOTMs is to improve learning by giving more weight to data deemed reliable or important. This approach offers a variable weight for each data instance.

To illustrate the effectiveness of these schemes, the authors apply automated and weighted SOTMs to two real-world datasets: country risk indicators to measure the evolution of global imbalances, and credit applicant data to assess the evolution of credit risks at the firm level.

\medskip
A new approach that incorporates limit cycles as an encoding mechanism to represent input patterns or sequences is described in \cite{DBLP:journals/nn/HuangGR15}. This method departs from the traditional static coding representations used in SOMs, where each input pattern is represented by a fixed point activation pattern on the map, a concept that is not consistent with the rhythmic oscillatory activity observed in the brain.

In this study, the authors develop and examine an alternative coding scheme that uses sparsely coded boundary cycles to represent external input patterns or sequences. They establish conditions under which limit cycle representations reliably learn and dominate the dynamics in a SOM. These limit cycles tend to be relatively unique for different inputs, robust to perturbations, and fairly insensitive to timing.

\subsection{Topology and Metrics }

\paragraph{Topology}

As already mentioned, SOMs are traditionally organized in square or hexagonal grids. Nevertheless, the selected topology has a great impact on the resulting SOM and has to be carefully considered.

\medskip
 In \cite{DBLP:journals/nn/Lopez-RubioR14}, the authors explore extensions of SOMs considering alternatives derived from the geometric theory of tessellations. The goal is to assess the effectiveness of these new topologies across various application domains such as unsupervised clustering, color image segmentation, and classification. Experimental findings indicate statistically significant differences between the topologies, suggesting that the optimal choice varies depending on the use case. This highlights the importance of customizing grid topology selection for each use case to maximize performance. Furthermore, the authors offer a theoretical interpretation of these results, providing insights into the underlying mechanisms that render certain topologies more efficient in particular contexts. This research sets the stage for further exploration of grid topologies in SOM applications, advocating for customization based on specific task requirements.
\medskip

In \cite{DBLP:journals/corr/AstudilloO15}, the authors introduce an improvement to self-organizing maps through the use of dynamic binary search tree (BST) structures. This approach, called TTOCONROT, allows dynamic adaptation and transformation of the SOM topology, thereby optimising data representation. The key innovation lies in the concept of 'Neural Promotion', where neurons adjust according to their increasing importance, more accurately reflecting the stochastic distribution of the data. Experimental results show that this method significantly improves the accuracy and dynamics of SOMs, without requiring the user to have an in-depth understanding of the topological properties of the data.

\medskip
"AMSOM: Adaptive Moving Self-organizing Map for Clustering and Visualization" \cite{DBLP:journals/corr/SpanakisW16} presents AMSOM that introduces significant flexibility by allowing dynamic adjustments to the position of neurons, as well as the addition and deletion of neurons during training, whereas traditional SOMs have a fixed structure during training. This approach addresses two major limitations of conventional SOMs: the rigidity of their neuron grid and the inability to delete neurons once they have been introduced. By offering a more adaptable structure, AMSOM not only improves training performance, but also the quality of data visualisation. Experiments carried out on various datasets have shown that AMSOM leads to a better representation of the input data and provides a useful framework for determining the optimal number of neurons and their structure, thereby improving the efficiency of clustering and visualisation.

\medskip
In \cite{DBLP:journals/neco/RougierD21}, the authors present an innovative variant of the self-organizing map algorithm by introducing the random placement of neurons on a two-dimensional surface. This approach, which follows a blue noise distribution, makes it possible to create various topologies with random but controllable discontinuities. This flexibility is particularly advantageous for organising high-dimensional data. The algorithm was tested on one-, two- and three-dimensional tasks, as well as on the MNIST handwritten digits dataset. The results, validated by spectral analysis and topological analysis of the data, show that this randomised self-organising map can effectively reorganise itself in the event of neuronal lesion or neurogenesis, demonstrating a remarkable capacity for adaptation. 

\medskip
DBGSOM (Directed Batch Growing Self-Organizing Map) \cite{DBLP:journals/asc/VasighiA17}  aims to optimise the quality of topology preservation when growing SOM maps. Unlike the usual incremental learning strategies of GSOMs, DBGSOM introduces a batch learning strategy that guides the growth process based on the accumulated error around the candidate neuron at the boundary. This method allows only one new neuron to be added around each candidate neuron, reducing the risk of map distortion and twisting, common problems in traditional GSOMs due to unexpected network growth and poor initialization of neuron weights.

\paragraph{Distances}

The impact of using different distance metrics on the cooperative process of the SOM algorithm is explored in \cite{DBLP:conf/isnn/Wilson17}. While the Euclidean distance is generally used in standard implementations of the algorithm, the study reveals that other metrics, such as the Manhattan distance and those from the same family as the Euclidean metric, can also converge and produce comparable results. However, the authors point out that simply being a metric is not enough to guarantee satisfactory results, and they  present examples of metrics that do not produce adequate maps despite their analogy with the Euclidean metric. This research sheds light on the need to choose distance metrics wisely in SOMs, taking into account their compatibility with the structure and objectives of the data being processed.

\medskip
A similarity measure called correntropy-induced metric (CIM) is explored in \cite{DBLP:journals/ijon/ChalasaniP15}. This approach aims to improve the magnification factor of the mapping, which is often distorted in standard SOMs due to the use of the mean square error, which tends to oversample regions of low probability. The study shows that adapting the SOM according to the CIM is equivalent to reducing the localised cross-information potential, an information theory function that quantifies the similarity between two probability distributions. By exploiting this property, the authors propose a kernel bandwidth adaptation algorithm for Gaussian kernels. The proposed model achieves a mapping with optimal magnification and automatically adapts the kernel function parameters, offering a significant improvement over traditional SOM approaches, without overcomplicating the algorithm.

\medskip
In \cite{DBLP:conf/ijcai/Celinska-Kopczynska22} a generalized framework for non-Euclidean SOMs is proposed. Traditionally, SOMs are based on Euclidean geometry, which limits their ability to model similarity relations in complex data. By adopting a non-Euclidean geometry, the authors open up new perspectives for dimension reduction, clustering and similarity discovery in large datasets. This innovative approach offers a new freedom in the translation of similarities into spatial neighbourhood relations, thus significantly improving the ability of SOMs to model complex and hierarchical data.

\medskip
Considering complex data such as images, graphs, matrices and time series, in \cite{DBLP:conf/dexa/DrakopoulosGMS20} the authors propose a multilinear approach to distances in the input vector space, offering greater flexibility in cluster formation. This methodological choice overcomes the traditional limits of fixed geometric shapes in clustering, opening up the possibility of creating clusters with arbitrary shapes.

The core of the study lies in the application of this distance metric to a multimodal functional neuroimaging (fMRI) dataset, taken during three distinct cognitive tasks. The aim was to assess the ability of SOMs to cluster this data in a meaningful way. The results obtained from various SOM configurations are evaluated using confusion matrices, topological error rates, activation set change rates, and intra-cluster distance variations.

\medskip
The idea of a bi-modal scaled metric is introduced in \cite{DBLP:journals/corr/abs-1805-03337}. This innovation allows more accurate segmentation of SOM maps into distinct regions, reflecting the expected cluster structure in the data. This is particularly relevant in the context of somatotopic maps, where clusters in the data may correspond to specific regions of the body surface. The use of this bi-scale metric helps to solve a common problem in SOMs: that of map neurons that are not activated by any training data. Thanks to this approach, the anticipated structure of the data is better preserved, and the SOM maps become more representative of the actual distribution of the data.

In addition, the study investigates the plasticity of SOM maps using this metric when they are re-trained following the loss of groups of neurons or following changes in the training data. These simulations are of particular importance for understanding how SOM maps can adapt to changes in the data, for example, in a neurobiological context where the loss of a region of the body surface requires the somatotopic mapping to be readjusted.

\paragraph{Neighborhood Relationships}

In \cite{DBLP:journals/tnn/HikawaM15}, a new neighborhood function specifically designed for hardware SOMs is proposed to enhance their vector quantization performance. This function, tailored to hardware constraints and utilizing negative powers of two, was tested through simulation and implemented on a Field-Programmable Gate Array (FPGA). The results demonstrate that this neighborhood function enhances the vector quantization performance of hardware SOMs without increasing hardware costs or slowing down operational speed.

\medskip
In \cite{DBLP:conf/icaisc/Olszewski14}, the SOM algorithm dynamically adjusts the neighbourhood width of neurons according to the frequency of occurrence of input patterns in the data space. This approach ensures that each neuron on the SOM grid has its own adapted neighbourhood width, enabling better visualisation of the data, especially when significant differences exist in the frequencies of the input patterns. The experimental study carried out on three real data sets confirms the effectiveness of this adaptive SOM approach, highlighting its potential for improving the visual representation and analysis of data.

\medskip
An innovative neighborhood function enabling continuous learning is introduced in \cite{DBLP:conf/ssci/HikawaIM18}. Unlike the traditional neighborhood function, whose intensity and radius decrease over time, this new function relies solely on the distance between the weight vector of the winning neuron and the input vector. This means that the magnitude and radius of the neighborhood function are adjusted based on this distance, rather than learning iterations. This approach allows the SOM to continue its learning uninterrupted, even in the presence of variable input distributions, which is particularly useful for online learning or in dynamic input environments. The use of vector distance in this method provides the SOM with a voluntary learning capability, akin to curiosity observed in the biological brain, enabling continuous adaptation to new information.

\medskip
\cite{DBLP:journals/eaai/Olszewski21} proposes to consider the scattering of input data to better preserve the main structure of the data. This method starts with a preliminary clustering of the input data to capture their scattering. Then, the intra-cluster variances obtained are used to define the neighborhood widths of the best matching units (BMUs). This approach was empirically evaluated on three diverse real-world datasets, varying in size, dimensionality, and type, representing different experimental domains. The performance of this method was compared to seven other data visualization techniques. The results demonstrated that the proposed method outperforms the other techniques in terms of efficiency, utility, and accuracy. This improved SOM method thus distinguishes itself by its ability to adapt more precisely to the structure of the data, thereby providing enhanced and more informative visualization.

\medskip
Using specific neighborhood relations\cite{DBLP:journals/npl/ElghazelB14} compares traditional techniques such as k-means and hierarchical clustering with innovative approaches based on graph coloring. The latter uses both dissimilarities and neighborhood relationships to partition the SOM, providing a more comprehensive analysis of the data. Experimental results demonstrate the effectiveness of these graphical methods in enhancing SOM clustering.

\medskip
 A generalisation of SOMs with 1-D neighbourhoods (chains of neurons) \cite{DBLP:journals/tnn/GorzalczanyR18} allows the network to split into sub-chains and dynamically regulate the total number of neurons. This feature gives the network the ability to automatically generate collections of multiprototypes capable of representing a wide range of clusters in datasets, all in a completely unsupervised manner. The paper also includes a sensitivity analysis to changes in control parameters and a comparative analysis with an alternative approach. These experiments demonstrate the effectiveness and adaptability of the proposed approach, offering a powerful method for cluster analysis in various data contexts.

\subsection{Learning Techniques}

Different learning approaches can optimise the effectiveness and efficiency of SOMs, focusing on aspects such as the rapid identification of optimal matching units, improved learning for weak or distant neurons, and the adaptation of learning techniques to the requirements of various applications.

\paragraph{General Learning Approach}

The learning efficiency of the least solicited or most distant neurons from the winning neuron is studied in \cite{DBLP:journals/ijfs/ChaudharyBA15}. In a standard SOM, these neurons receive less exposure to input data, reducing their learning capacity. To avoid this, the authors propose CSOM, a technique that enhances the learning of these so-called "weak" or "distant" neurons. This approach aims to strengthen learning within the community of winning neurons, which could lead to better knowledge distribution and more precise mapping of input data. By increasing the efficiency of learning in less active neurons, CSOM could potentially improve the overall performance of SOM in various applications, including classification, data visualization, and clustering.

\medskip
In \cite{DBLP:conf/icann/BernardHG20}, the authors tackle the challenge of computational efficiency in the learning phase, especially when a large number of neurons is involved. Finding the Best Matching Unit (BMU) is a key process in SOM operation, but it can become increasingly time-consuming as the number of neurons increases. The algorithm proposed in this paper aims to significantly accelerate the search for the BMU while minimizing performance loss. This speed improvement could make SOMs more practical and efficient for applications requiring a large number of neurons, such as complex pattern recognition or voluminous data analysis. The proposed approach thus paves the way for more ambitious uses of SOMs, where the size and complexity of models are no longer major constraints.

\medskip
An approach, called ‘input information maximisation’ \cite{DBLP:journals/apin/Kamimura14}, focuses on the input neurons by considering mainly the winning neurons. The method is based on evaluating the uncertainty of the input neurons, defined by the difference between the input neurons and the corresponding winning neurons, and then normalised to obtain a clear measure.

The central idea is that increasing the input information leads to reduced activation of the input neurons, with the maximum state characterised by the activation of a single neuron and the deactivation of all the others. This property is particularly useful for reducing quantization and topographical errors in SOMs, thus improving the overall quality of the representations obtained.

The authors applied this method to two distinct datasets: Senate voting data and voting attitudes. The experimental results confirmed that increasing the input information leads to a reduction in quantification and topographical errors, while allowing a clearer class structure to be extracted.

\medskip
"Two novel hybrid Self-Organizing Map based emotional learning algorithms" (EmSOM/Em-SOR-SOM)\cite{DBLP:journals/nca/DaiG19} introduces two innovative algorithms, EmSOM and Em-SOR-SOM, for integrating emotional aspects into machine learning. These algorithms aim to improve the modelling of human reactions by taking account of emotions, an essential but often neglected aspect of decision-making.

EmSOM corrects the shortcomings of Emotional Backpropagation (EmBP) by determining emotional input values from the corresponding SOM blocks. This approach provides a more accurate and nuanced representation of emotions in the learning process.

Em-SOR-SOM combines the SOR-SOM (Sparse Online SOM) algorithm with emotional learning to improve pattern recognition. This hybridisation exploits the advantages of SOR-SOM, in particular its ability to manage sparse data efficiently.

The performance of EmSOM and Em-SOR-SOM has been validated on facial recognition and credit databases. These algorithms have demonstrated their superiority over EmBP and other recent methods, paving the way for advanced applications in pattern recognition and automated decision-making systems.

\paragraph{Features Extraction}

"A Convolutional Deep Self-Organizing Map Feature Extraction for Machine Learning" \cite{DBLP:journals/mta/SakkariZ20} introduces a novel approach to feature extraction called Unsupervised Deep Self-Organizing Map (UDSOM). This method merges multi-layer SOM architecture with deep learning principles to tackle challenges associated with high-dimensional data handling and feature extraction in Big Data environments.

UDSOM's key innovation lies in its data processing workflow, which involves segmenting data into sub-regions, applying self-organizing layers, and rectification functions (RELU). Each SOM layer focuses on modeling a local sub-region, and the most active neurons from each SOM are grouped in a sampling layer to generate a new 2D map. Concurrently, data abstraction occurs through a convolution-pooling module and ReLU function application. This architecture facilitates the collection and integration of local information to construct a global representation in upper layers, crucial for effectively handling Big Data. Experimental results demonstrate the effectiveness of UDSOM in various machine learning tasks, highlighting its potential for feature extraction in the context of large-scale datasets.

\medskip

An innovative approach \cite{DBLP:conf/iconip/KhacefRM20} focuses on exploiting extracted features. This method is particularly relevant in the context of embedded applications, where the efficiency and accuracy of unsupervised learning are paramount.

The core of this study lies in the comparison of two feature extraction methods: one based on machine learning with Parsimonious Convolutional Autoencoders and the other inspired by neuroscience with Pulse Neural Networks using pulse timing-dependent learning. This contrast between a traditional machine learning approach and a more experimental neuroscience approach offers a fascinating insight into how data features can be effectively extracted and used to improve SOM classification.

\paragraph{Training Phase}

"VSOM: Efficient, Stochastic Self-Organizing Map Training" \cite{DBLP:conf/intellisys/Hamel18} presents VSOM, an efficient implementation for stochastic training of Self-Organizing Maps. VSOM improves on Kohonen's standard stochastic algorithm by replacing iterative constructions with vector and matrix operations. This innovative approach delivers significant performance gains over Kohonen's iterative algorithm and over batchSOM, the fastest implementation of SOM without the need for multi-processing.

The quality of the maps produced by VSOM is comparable to that obtained with the original iterative algorithm and surpasses that of batchSOM. Suitable for single-threaded operation, VSOM lends itself particularly well as a replacement for iterative stochastic SOM training, especially in environments such as R that do not handle multi-threading efficiently.

\medskip
In \cite{DBLP:journals/ijssmet/TatoianH18}, the authors introduce a new quality index called the convergence index. This index combines map fitting accuracy and estimated topographic accuracy, thus providing a single, statistically significant number that proves to be more intuitive to use than other quality measures. This research focuses on how the convergence index captures SOMs learning the multivariate distribution of a training dataset. Special attention is paid to the convergence of the marginals and the influence of different parameters governing SOM learning. Surprisingly, it is found that the constant neighborhood function outperforms the popular Gaussian neighborhood function in producing more effective SOM models. This result highlights the importance of choosing an appropriate neighborhood function to optimize the learning and convergence of self-organizing maps.

\paragraph{Clustering}

In \cite{DBLP:conf/iconip/MaurelSL17}, the authors address the concept of collaborative clustering, a technique aimed at revealing the common structures of data spread over different sites. In today's environment, where the amount of data available is constantly increasing, incremental clustering methods are essential. The algorithm presented in this article enables collaborative clustering to be carried out using Self-Organizing Maps in an incremental manner, without requiring any topological modifications to the map. This approach is particularly useful for efficiently managing large volumes of data that are constantly changing.

Experiments carried out on several datasets validate the proposed method. The article also highlights the influence of batch size on the learning process, a crucial aspect for optimising clustering performance in dynamic and collaborative environments. This work thus represents a significant contribution both to the field of incremental clustering and to that of SOMs applied to collaborative clustering.

\medskip
\cite{DBLP:journals/igpl/PasaCM17} explores an innovative approach in data clustering using SOMs. This research focuses on improving the generalisation and performance of SOMs using an ensemble method.

Data clustering is an analytical method for identifying groups within data based on similarities. However, clustering results can vary depending on many factors such as algorithm parameters, initialization, stopping criteria, or the use of different attributes or data subsets. SOMs are frequently used for data analysis, particularly for visualisation and clustering.

The approach proposed in this article is to use a set of SOM networks working independently, combined by a system that integrates the individual results into a single output. This method aims to achieve better generalisation than using a single neural network. The key concept of this method is the use of cluster validity indices to combine the weights of neurons from maps of different sizes.

\medskip
\cite{DBLP:conf/hais/PasaCM15} proposes an innovative method for improving classification accuracy by fusing SOMs of different sizes. This approach, based on the neural network ensemble, aims to obtain a better generalisation of the model. Through factorial experiments and computer simulations using datasets from the UCI Machine Learning Repository and the Fundamental Clustering Problems Suite, the authors demonstrate a significant increase in classification accuracy. The Wilcoxon Signed Rank test validates the feasibility of this method, highlighting its potential for improving SOM-based classification techniques.

\medskip
In "Improving self organizing maps method for data clustering and classification" \cite{DBLP:conf/ascc/ZinYM15}, iSOM bis is presented as an improvement of Self-Organizing Maps to solve visualization and classification difficulties. This model improves the distinction between clusters and sub-clusters by adding a third dimension based on computational distance to the winning neurons. Tested on the Iris Flowers dataset, iSOM showed improved class separation and increased classification accuracy compared with traditional SOMs, offering a more efficient method for analysing complex data.

\subsection{Visualization}

\medskip
"A novel data-driven visualization of n-dimensional feasible region using interpretable self-organizing maps (iSOM)" \cite{DBLP:journals/nn/NagarPR22} addresses the complexity of visualising optimisation problems in multidimensional spaces. Traditionally, graphical optimisation has been limited to one- or two-dimensional problems, where visualisation of the contours of objective functions and constraints facilitates understanding and decision-making. However, this method reaches its limits as the dimensions increase.

To overcome this difficulty, the authors introduce the use of iSOM, a variant of artificial neural networks, for the visualisation of multidimensional data. Thanks to iSOM, it is possible to graphically represent regions that can be realised in n-dimensions on two-dimensional representations. This method is embodied by the creation of a "B-matrix", a graphical representation that visualises both the feasible range of design variables and the objective function.

B-matrix effectively reduces the complexity inherent in multi-dimensional search spaces, enabling more intuitive data analysis. The flexibility and effectiveness of this method is demonstrated through various analytical and engineering examples, covering dimensions from 2 to 30. This approach represents a significant advance for designers and engineers, offering a new way of approaching and solving complex optimisation problems in many fields.

\medskip
In \cite{DBLP:conf/mmm/PeskaL22}, a modification of the SOM algorithm incorporates relevance scores into the 2D data visualization process. Traditionally, SOMs are effective at preserving the topological relationships of multidimensional data, but they do not take into account the relevance of the results, a crucial aspect in multimedia information retrieval systems. The proposed modification aims to simultaneously optimise the preservation of topological order and result relevance. This approach not only improves data exploration capability, but also enables the most relevant results to be exploited more effectively, providing a richer, contextually relevant visualisation.

\medskip
"Cluster and visualize data using 3D self-organizing maps" \cite{DBLP:conf/urai/Zin14} presents an extension named 3D-SOM that  improves the clustering and visualisation capabilities of multidimensional data. 3D-SOM develops SOM algorithms in terms of the number, relationship and structural arrangement of output neurons, as well as neighbourhood weight updating processes and distance calculations in three-dimensional space.

Tested on the 'Iris Flowers' dataset, 3D-SOM showed a richer representation of the data in three dimensions and significantly reduced quantification and topographical errors compared with conventional SOM, demonstrating its effectiveness in cluster formation and data visualisation.

\medskip
In \cite{DBLP:journals/corr/abs-1910-01590}, the authors address an important issue in the field of complex data visualization and processing: the ability to generate interpretable visualizations while performing efficient clustering. The authors propose an innovative fusion of clustering and representation learning techniques. Their approach introduces a new method called PSOM for adapting SOMs with probabilistic cluster assignments. This advance makes it possible to combine the visual advantages of SOMs with improved clustering performance, a notable shortcoming in existing methods.

In addition, the paper proposes the DPSOM architecture, an integration of the SOM with a variational auto-encoder (VAE), for probabilistic clustering. This combination offers superior clustering performance compared to current deep clustering methods, while preserving the beneficial visualization properties of SOMs. The T-DPSOM extension for time series clustering is particularly noteworthy, as it enables prediction in latent space using LSTM networks, providing interpretable visualisations of patient state trajectories and uncertainty estimation.

\medskip
"Landmark map: An extension of the self-organizing map for a user-intended nonlinear projection" \cite{DBLP:journals/ijon/Onishi20} presents an innovation in the field of SOMs, responding to the problem of personalised data visualisation. Traditional SOMs, while effective for dimensionality reduction, often lack the flexibility to adapt their projections to users' specific needs. LAMA, by incorporating landmarks, allows users to directly influence the projection of data, leading to more intuitive, user-centred visualisations. This approach is particularly useful in contexts where the relationship between data and its spatial representation is crucial, such as in recommendation systems or human-computer interaction. By enabling non-linear projection in line with the user's intentions, LAMA opens up new possibilities for more targeted and meaningful data exploration.

\subsection{Performances}

The computational cost of SOMs algorithms being high, different improvement paths can be explored to enhance the map generation process.

\medskip
"New Hardware Architecture for Self-Organizing Map Used for Color Vector Quantization"  (D-SOM) \cite{DBLP:journals/jcsc/KhalifaBAB20} tackles the problem of optimising the performance of Self-Organizing Maps by proposing a new hardware architecture. The Diagonal-SOM (D-SOM) is presented as an innovative solution for accelerating SOM processing by exploiting two-level parallelism between neurons and connections. This approach aims to improve the efficiency of data processing, particularly in the context of vector colour quantisation.

The D-SOM, designed as a core Hardware-Description-Language, is characterised by its flexibility and adaptability to various SOM topologies. Validation of this architecture has demonstrated temporal performance that is almost twice as good as that obtained in recent studies. This advance represents a significant step forward in the development of faster and more efficient SOM architectures for real-time data processing applications.

\medskip
"AW-SOM, an Algorithm for High-Speed Learning in Hardware Self-Organizing Maps" \cite{DBLP:journals/tcas/CardarilliNFRS20} tackles a crucial limitation of hardware implementations of Self-Organizing Maps: the reduction in processing speed as the number of neurons and features increases. SOMs are widely used for clustering, but their performance can be hampered by processing speed constraints, particularly in hardware architectures such as Field Programmable Gate Arrays (FPGAs).

To address this problem, the authors introduce All Winner-SOM (AW-SOM), a modified version of the traditional SOM algorithm. The main advantage of AW-SOM is its ability to maintain a processing speed that is almost independent of the number of neurons, reaching a clock frequency of 145 MHz for configurations ranging from 16 to 256 neurons. This feature represents a breakthrough for hardware SOM applications, enabling a significant increase in system performance in terms of computation rate.

\medskip
In \cite{DBLP:conf/ausai/NguyenHT16}, the authors introduce a breakthrough in the implementation of SOMs, harnessing the parallel computing power of graphics processing units (GPUs) to substantially increase the resolution of the SOM's display space. This enhancement makes it possible to visualise high-dimensional data on a high-resolution display space in a reasonable computing time.

The main advantage of this approach is the ability of high-resolution SOM to reveal complex details about relationships between input feature vectors that would otherwise be lost with low-resolution SOM. This is demonstrated through an application to an artificially generated dataset, the ‘policeman’ dataset, designed to illustrate complex relationships between feature vectors. Using the power of GPUs to increase the resolution of the SOM opens up new possibilities for the exploration and detailed understanding of data in various application domains. 

\medskip
\cite{DBLP:journals/nca/GirauT20} explores the fault tolerance of self-organizing maps, focusing on targeted implementations on Field Programmable Gate Arrays (FPGAs). The study uses a bit-reversal fault model to inject errors into registers, assessing the impact on SOM performance through quantisation and distortion measurements on synthetic datasets. The paper examines three passive techniques to improve SOM fault tolerance during learning. Experimental results reveal an important property of SOMs: smooth degradation in the face of faults. Furthermore, the fault tolerance of SOMs can be significantly improved depending on certain technological choices, such as sequential or parallel implementation and weight storage policies. The authors show that SOMs can become highly fault-tolerant, especially when weights are stored individually in the implementation.

%%%%%%%%%%%%%%%%%%%%%%%%%%%%%%
%%%%%%%%%%%%%%%%%%%%%%%%%%%%%%%
%%%%%%%%%%%%%%%%%%%%%%%%%%%%%%

\subsection{Hyperparameterization}
 
To achieve better results, hyperparameter tuning for Self-Organizing Map algorithms has long been recognized as crucial \cite{Utsugi97}, yet it remains underexplored. More broadly, automated configuration techniques, encapsulated in the concept of AutoML, are now prevalent in machine learning \cite{PoulakisDK24}. This core challenge is rooted in the foundational problem of algorithm selection \cite{Smith-Miles08}, which has been extensively studied for optimizing algorithms \cite{Hoos12}. As a result, highly efficient automated parameter configuration systems are now available. 
 Unlike model parameters, which are learned automatically during training, hyperparameters are defined before the learning process and remain constant during it.

In the context of SOMs, hyperparameters can include the selection of the grid size, the choice of neighbourhood function, the learning rate, the number of iterations, and other crucial aspects that determine the quality of the mapping and the speed of convergence of the network. The appropriate setting of these hyperparameters is essential to obtain optimal results, as an inappropriate choice can lead to slow convergence, inaccurate representation of the data or even training failure.

Hyperparameterisation is often a challenge, as it requires a thorough understanding of how each hyperparameter affects the model and a careful exploration of the hyperparameter space. Methods such as cross-validation, grid search and more sophisticated techniques such as Bayesian optimisation are used to find an optimal set of hyperparameters. The aim is to improve model performance while maintaining manageable complexity and computation time.

\medskip
In \cite{DBLP:journals/ijcopi/Jaramillo-Vacio15}, the authors use the response surface methodology (RSM) and the desirability function, a combination that makes it possible to determine the optimal conditions for improving SOM performance.The analysis focuses on the relationship between various explanatory variables, such as the competitive algorithm and learning rate, and response variables such as training time and quality metrics for the SOM. Using response surface plots, the authors examine the interaction effects of the main factors and determine the optimal conditions for partial discharge classification performance.

This methodological approach reveals valuable insights into how different parameters influence the overall performance of the SOM, providing strategic guidance for fine-tuning parameters in specific applications. The use of the desirability function, in particular, adds an interesting new dimension, helping to balance and optimise multiple performance objectives simultaneously, which is crucial in complex contexts such as partial discharge classification.

\paragraph{Convergence}

One of the major challenges in using SOMs is indeed the efficient and reliable convergence of the learning algorithm.

\medskip

\cite{DBLP:journals/ccsecis/KuoC16} introduces the concept of momentum into the learning process. This modification aims to overcome one of the fundamental problems of SOMs, namely their tendency to become trapped in local optima. By incorporating momentum, similar to what is used in the backpropagation algorithm for neural networks, the author seeks to give neurons the ability to ‘jump’ beyond these local optima, thus offering the potential to improve network convergence.

The study also addresses the issue of momentum management, taking into account the previous movement of neurons. This consideration is essential because it has a direct influence on the learning dynamics of the network, particularly for neurons that are not immediate winners but are in the neighbourhood of the winning neuron. This finesse in momentum management is a response to the specific nature of SOMs, where the movement of a neuron is highly dependent on its current role in the network (winning or non-winning).

\medskip

In \cite{DBLP:journals/cogcom/KhanXWQHV15}, the authors tackle the theoretical understanding and practical application of self-organizing maps in the field of derivative-free optimization. The SOMO algorithm, an innovative variant of the classical SOM, is designed to solve continuous optimisation problems, a domain where SOMs have traditionally not been widely used.

One of the major contributions of this work is the establishment of formal convergence proofs for the SOMO algorithm. These proofs, based on a specific distance measure, guarantee that the distance between neurons decreases with each iteration, eventually converging to zero. In addition, it is shown that the value of the function of the winning neuron decreases after each iteration, an important point for the efficiency of the optimisation.

\paragraph{Parameters}

Different strategies and advanced methodologies have been devised for tuning and optimising SOMs parameters to improve the convergence, accuracy and overall efficiency in processing and visualising complex data.

\medskip

In \cite{DBLP:conf/aisi/AhmedSM19}, the authors point out that the quality and accuracy of SOM results strongly depend on the choice of hyperparameters such as map size, number of iterations, and initial learning rate.

The paper focuses on the use of a genetic algorithm to optimise these parameters. A genetic algorithm is a biologically inspired search method that simulates the process of natural selection to identify the best solutions. The authors propose a new approach for selecting and optimising SOM hyperparameters using this technique. They present their first results by applying this method to the classical colour clustering problem. This application serves as a case study to demonstrate the effectiveness of their proposal. The choice of the colour clustering problem provides a solid basis for evaluating the performance of the genetic algorithm in optimising SOM parameters, as it is a well-known and relatively simple problem.

\medskip

\cite{DBLP:conf/iconip/Saini0MB20} presents a neural network-based unsupervised classification technique for tweet summarization, selecting informative tweets based on their importance. The method is implemented in two steps: in the first step, a self-organising map is used to reduce the number of tweets. In the second step, a granular self-organising map (GSOM), which is a two-layer neural network using fuzzy and rough set theory for training, is used to cluster the reduced set of tweets. A fixed-length summary is then generated by selecting tweets from the resulting clusters. GSOM has a set of parameters whose proper selection influences performance. Therefore, an evolutionary optimisation technique is used to select the best combinations of parameters. The effectiveness of the proposed approach has been evaluated on four disaster-related microblog datasets. The results clearly show that the proposed method outperforms the most advanced current methods.

\medskip

‘A model to estimate the Self-Organizing Maps grid dimension for Prototype Generation’ \cite{DBLP:journals/ida/SilvaVD21} discusses the improvement of the K-nearest neighbour (KNN) classifier using a prototype generation (PG) method to reduce classification time, particularly for large datasets. KNN is accurate but slow, and PG aims to represent data examples by prototypes, making classification faster. However, determining the ideal number of prototypes remains a challenge. This study proposes a model to estimate the best grid size for self-organising maps and thus determine the optimal number of prototypes based on the number of examples in the dataset. Compared with other PG methods using artificial intelligence, the proposed method has the advantage of maintaining a good balance between a reduced number of prototypes and accuracy, without degrading the classification performance of KNN.

\medskip
"A constant learning rate self-organizing map (CLRSOM) learning algorithm" \cite{DBLP:journals/jise/ChaudharyBA15} proposes to use of a constant learning rate. Unlike traditional SOMs, which require a precisely calibrated decreasing learning rate, CLRSOM maintains a constant learning rate throughout the process. This method simplifies the choice of learning rate, a common challenge with traditional SOMs where an unsuitable rate can lead to insufficient convergence or excessive processing time.

CLRSOM intelligently selects both the neuron closest to and furthest from the Best Match Unit (BMU), delivering improved performance despite a constant learning rate. Applied to various standard datasets, CLRSOM demonstrated a substantial improvement in learning performance, preserving the topology of the input data without sacrificing quantification error and neuron utilisation levels, compared with traditional SOMs.

\medskip
\cite{DBLP:journals/remotesensing/KristollariK20} focuses on the use of SOM for cloud masking in Sentinel-2 satellite images, a promising method thanks to its ability to preserve topological relationships and its faster training time compared to other neural networks. The authors propose a fine-tuning methodology to improve the performance of the SOM in separating clouds from bright land surfaces. This method is based on the adjustment of incorrect neural labels identified by the initial network, without the need for additional training.

The study is notable for its methodical approach: the SOM network was trained on the largest public spectral database for cloud masking in Sentinel-2 and tested on an independent database. The evaluation of the fine-tuned SOM was carried out both qualitatively and quantitatively, using several visualisation techniques to interpret its behaviour.

Table \ref{table:biblio} provides an overview of the previously cited works according to our analysis scheme depicted in Figure \ref{fig:orga_general}. 

\begin{tiny}
\begin{table}
\begin{tabular}{|c|l|l|}
\hline
Data management & Semi-supervized Datasets& \cite{DBLP:conf/ijcnn/BragaB18,DBLP:journals/ijon/AllahyarYH15}\\

		& Disturbed Datasets & \cite{DBLP:conf/icnc/StarkeyA17,DBLP:journals/simpa/GouveaGMC23}\\

		& Categorical Data & \cite{DBLP:journals/ida/HsuKJC19,DBLP:journals/asc/CosoFDNRV15,DBLP:conf/iv/KilgoreTCN20}\\

		& Particular Data Distributions & \cite{DBLP:journals/classification/CarvalhoIVB22,DBLP:journals/isci/DUrsoGM20}\\

		& Large Datasets & \cite{DBLP:journals/ijcia/CordelA18,DBLP:conf/ijcnn/AntoninoA16}\\

		& Time and Patterns & \cite{DBLP:journals/kais/Sarlin15,DBLP:journals/nn/HuangGR15}\\
\hline 
Topology and Metrics & Topology & \cite{DBLP:journals/nn/Lopez-RubioR14,DBLP:journals/corr/AstudilloO15,DBLP:journals/corr/SpanakisW16,DBLP:journals/neco/RougierD21,DBLP:journals/asc/VasighiA17}\\

		& Distances & \cite{DBLP:conf/isnn/Wilson17,DBLP:journals/ijon/ChalasaniP15,DBLP:conf/ijcai/Celinska-Kopczynska22,DBLP:conf/dexa/DrakopoulosGMS20,DBLP:journals/corr/abs-1805-03337}\\

		& Neighborhood Relationships & \cite{DBLP:journals/tnn/HikawaM15,DBLP:conf/icaisc/Olszewski14,DBLP:conf/ssci/HikawaIM18,DBLP:journals/eaai/Olszewski21,DBLP:journals/npl/ElghazelB14,DBLP:journals/tnn/GorzalczanyR18}\\
\hline
Learning Techniques & General Learning Approach & \cite{DBLP:journals/ijfs/ChaudharyBA15,DBLP:conf/icann/BernardHG20,DBLP:journals/apin/Kamimura14,DBLP:journals/nca/DaiG19}\\

		& Features Extraction & \cite{DBLP:journals/mta/SakkariZ20,DBLP:conf/iconip/KhacefRM20} \\

		& Training Phase & \cite{DBLP:conf/intellisys/Hamel18,DBLP:journals/ijssmet/TatoianH18}\\

		& Clustering & \cite{DBLP:conf/iconip/MaurelSL17,DBLP:journals/igpl/PasaCM17,DBLP:conf/hais/PasaCM15,DBLP:conf/ascc/ZinYM15}\\
\hline
Visualization & & \cite{DBLP:journals/nn/NagarPR22,DBLP:conf/mmm/PeskaL22,DBLP:conf/urai/Zin14,DBLP:journals/corr/abs-1910-01590,DBLP:journals/ijon/Onishi20}\\
\hline
Performances & & \cite{DBLP:journals/jcsc/KhalifaBAB20,DBLP:journals/tcas/CardarilliNFRS20,DBLP:conf/ausai/NguyenHT16,DBLP:journals/nca/GirauT20}\\
\hline
Hyperparameterization & Convergence & \cite{DBLP:journals/ccsecis/KuoC16,DBLP:journals/cogcom/KhanXWQHV15}\\

		& Parameters & \cite{DBLP:conf/aisi/AhmedSM19,DBLP:conf/iconip/Saini0MB20,DBLP:journals/ida/SilvaVD21,DBLP:journals/remotesensing/KristollariK20}\\
\hline
\end{tabular}

\caption{Bibliography summary of previously cited works}
\label{table:biblio}
\end{table}
\end{tiny}

\section{Applications of SOM : Focus on Customers Data}

The application of self-organising maps in specific areas, notably marketing and customer-focused products, has considerable potential, although it is still relatively unexploited compared with other sectors such as medical applications. SOMs, known for their effective clustering and visualisation of complex data, can offer unique insights into the analysis of consumer behaviour, customer segmentation, and the development of personalised recommendations. In this section, we review how SOMs can be used to better understand and respond to customer needs and preferences. The following studies illustrate the potential of SOMs to provide innovative solutions in marketing and product development, in sectors where understanding customer preferences and behaviours is crucial.

\medskip

‘Online recommendation based on incremental-input self-organizing map’ \cite{DBLP:journals/ecra/ZhouTL21} addresses a crucial issue in e-commerce: improving online recommendation models. In e-commerce platforms, recommendation algorithms play a key role in helping consumers discover products that may be of interest to them. However, most traditional recommendation models, based on static data, run into difficulties when it comes to managing new consumer reviews or recently added products.

The main drawback of static models is the need to re-train them offline with new evaluations, a process that is often inefficient and time-consuming. To solve this problem, this paper proposes an innovative online recommendation model using a self-organising map with incremental entries. This approach allows new consumers to be added to the model by increasing the input units and updating the corresponding weights between the input layer and the output layer. The new products are treated as new samples arriving at the SOM, and an incremental learning strategy specifically designed for the SOM is implemented.

Thanks to this method, the proposed model does not need to be retrained and can be updated online with very few evaluations of new consumers or products. Experimental results on real datasets show that the proposed algorithm outperforms several existing online recommendation algorithms in terms of recommendation accuracy, and offers similar performance to the static recommendation algorithm which requires retraining with the dataset. By integrating real-time data and continually adapting the model to new users and products, this algorithm can dramatically improve the online shopping experience, making recommendations more relevant and personalized.

\medskip
 An innovative approach to customer segmentation in the online retail business is proposed in \cite{DBLP:conf/icic/VohraPHGL20}. The RFM (Recency, Frequency, Amount) model is an analytical method used to segment customers based on their purchase history. It evaluates customers on the basis of three criteria: recency (when a customer last made a purchase), frequency (how often a customer makes purchases) and monetary amount (how much a customer spends).

In this paper, the performance of k-means clustering algorithms and SOMs are compared for the filtered target dataset. SOMs are used to provide a neural network computational framework, which can be compared to the simple k-means algorithm used by Chen et al. This hybrid approach, combining SOMs and k-means, offers a new perspective in the analysis of customer shopping behaviour and segmentation in the online retail sector.

The use of SOM, in particular, enables a more intuitive visualisation of customer groups and can reveal complex patterns in the data that are not immediately apparent with traditional methods such as k-means. This method therefore offers an opportunity to improve the personalisation of marketing and sales strategies, by targeting different customer segments more effectively according to their specific buying behaviour.

\medskip

An emerging business paradigm: the smart product-service system (Smart PSS) is studied in \cite{DBLP:journals/aei/CongCMXD23}. This concept combines smart and connected products (SCPs) with associated services, offering a set of integrated solutions. The study recognises the growing importance of Smart PSS in research, particularly in terms of conceptual design that considers both PCS and services simultaneously.

A major challenge identified in the design of Smart PSS is matching the proposed solutions with users' emotional requirements. To address this challenge, the paper proposes a conceptual design method based on the analysis of emotions generated by users during their interactions with products. This method involves identifying relevant traditional products, analysing user emotions from public opinion data, and using an interactive emotion map as a design tool to organise these emotions and associated design elements.

SOM plays a key role in this method, as it is used for grouping product samples and Kansei words (sensations or emotions in Japanese), thus facilitating the analysis and organisation of emotional data. In addition, the process of evaluating the improved solution is carried out using Hierarchical Process Analysis (HPA), a structured method for organising and analysing complex decisions.

\medskip

In \cite{DBLP:conf/iwost/PhamT15}, the authors introduce an innovative method for evaluating the quality of bio-food products in the context of global markets. This approach focuses particularly on cross-cultural consumer behaviour and their inclination towards healthy food.

The proposed method uses Kansei evaluation, which is a Japanese technique for capturing and analysing consumers' emotional reactions to a product or service. This evaluation is integrated with fuzzy rules and a self-organising map model to better understand and respond to the preferences and sensitivities of customers and experts. The aim is to select appropriate bio-food product and food technology alternatives that correspond to the behaviours of mass consumer and expert groups. The model aims to improve the quality of food products in global bio-food markets by taking into account diverse cultural preferences and healthy consumption trends. To validate the effectiveness of this approach, the model has been experimentally tested and applied in several areas in Asian countries.

%\paragraph{Conclusion}

%Advanced methods such as the integration of SOMs with deep neural network architectures and the exploration of unsupervised learning techniques for feature extraction have demonstrated a significant improvement in the classification and visualisation capabilities of SOMs. Similarly, the application of fine-tuning methods and the use of ensemble strategies have shown increased effectiveness in specific contexts, such as separating clouds from bright surfaces in satellite images or improving the classification of partial discharges.

\section{Conclusion}

%Recent improvements have focused on the ability of SOMs to effectively manage a variety of data types, whether categorical, numerical or mixed, while incorporating aspects such as the management of missing data or the visualisation of large masses of data. This phase of innovation in the SOM field reveals not only the versatility of this technique, but also its growing potential to meet the complex needs of modern data analysis. With the increasing complexity and diversity of data, the flexibility of SOMs has become essential. Hence, there is a need to tackle specific data challenges, such as managing missing data, integrating various data types (numeric and categorical), and adapting to complex data structures.

Concluding this survey, it is clear that there are many variations of the Self-Organizing Maps (SOM) algorithm, each seeking to improve or adapt Kohonen’s original method to specific challenges. From adjusting learning rates in the CLRSOM, to incorporating additional dimensions in the 3D-SOM, to introducing automatic feature weights in the FWSOM, these developments illustrate the versatility and continued evolution of SOMs. These variations reflect not only technological advances in the field of machine learning, but also a deeper understanding of the specific analytical needs of different datasets.

Recent improvements have focused on the ability of SOMs to effectively manage a variety of data types, whether categorical, numerical or mixed, while incorporating aspects such as the management of missing data or the visualisation of large masses of data. This phase of innovation in the SOM field reveals its growing potential to meet the complex needs of modern data analysis. With the increasing complexity and diversity of data, the flexibility of SOMs has become essential. Hence, there is a need to tackle specific data challenges, such as managing missing data, integrating various data types (numeric and categorical), and adapting to complex data structures.

\newcommand{\etalchar}[1]{$^{#1}$}

\end{document}